%% file: iclr2027_conference.tex
\documentclass{article} 
\usepackage{iclr2027_conference,times}

\input{math_commands.tex}

\usepackage{hyperref}
\usepackage{url}

\usepackage{amsmath}
\usepackage{amssymb}
\usepackage{mathtools}
\usepackage{amsthm}
\usepackage{enumitem}
\usepackage{algorithm}
\usepackage{algpseudocode}
\usepackage{array}
\usepackage{booktabs}
\usepackage{wrapstuff}
\usepackage{wrapfig}
\usepackage{lipsum}
\usepackage{graphicx}
\usepackage{caption}
\usepackage{capt-of}
\usepackage{placeins}
\title{Scalable In-Context Reinforcement Learning with Recurrent Algorithm Distillation}

\author{Yuanqing Ma, Zhenrui Zheng, Chenjun Xiao\\
The Chinese University of Hong Kong, Shenzhen\\
\texttt{\{yuanqingma, zhenruizheng1\}@link.cuhk.edu.cn}, \texttt{chenjunx@cuhk.edu.cn} 
}

\iclrfinalcopy 
\begin{document}

\maketitle

\begin{abstract}
Algorithm Distillation (AD) has demonstrated the remarkable ability of Transformers to perform in-context reinforcement learning  without explicit weight updates. 
However, capturing long-term learning progress necessitates expansive context windows, which incur prohibitive memory costs and limit scalability in complex, long-horizon tasks. 
To address this bottleneck, we propose \emph{Recurrent Algorithm Distillation (RAD)}. 
RAD employs a dual-component architecture: a \emph{Compression Transformer} that distills extended interaction histories into compact latent tokens, and an \emph{AD Transformer} that auto-regressively generates actions using a hybrid context of these compressed memories and recent transitions.
By maintaining a fixed-size latent buffer, RAD decouples the effective history length from computational complexity, functionally providing the model with a long-horizon memory. 
Empirical evaluations across diverse environments demonstrate that RAD matches the asymptotic performance of standard AD with significantly reduced context window sizes, offering a scalable solution for efficient in-context decision-making. Code is available at \url{https://github.com/tommyma3/rad}.
\end{abstract}

\begin{figure}[h]
    \centering
    \includegraphics[width=0.99\linewidth]{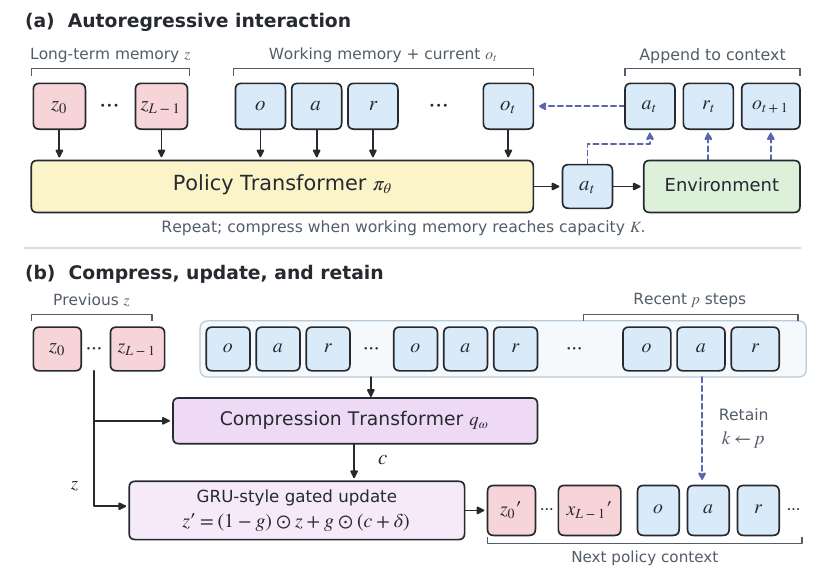}
    \caption{\textbf{RAD inference workflow:} During evaluation, the AD Transformer $\pi_\theta$ is able to autoregressively rollout the policy by predicting actions conditioned on the long-term memory $z$ and the working memory $x_{t-k:t}$. When working memory reaches its capacity $K$, the Compression Transformer \(q_\omega\) compresses the previous long-term memory (represented as $L$ latent tokens) and the older working-memory context into candidate latent tokens. The update gate \(U_\phi\) integrates these candidates with the previous latent state to form the new long-term memory \(z'\), while the most recent \(p\) steps remain in working memory. At the first compression, since no previous long-term memory exists, \(z\) is set directly to $q_\omega(x_{0:K})$, bypassing the update gate.}
    \label{fig:RADinference}
\end{figure}

\section{Introduction}

In-Context Reinforcement Learning (ICRL) enables agents to adapt to new tasks during inference without parameter updates. A prominent example is Algorithm Distillation (AD) \citep{laskin2022incontextreinforcementlearningalgorithm}, which frames RL as a sequence modeling problem. By training on across-episodic learning histories of source RL algorithms, AD distills the ``improvement operator'' itself, enabling agents to autoregressively predict actions and solve unseen tasks by attending to their own interaction history.
Despite its promise, to capture the ``improvement operator'', AD requires an expansive context window to encompass sufficient transitions, which is expensive to process due to Transformers' quadratic complexity \citep{vaswani2017attention}. Pratical implementations therefore truncate older history, which may potentially discard information essential for long-horizon adaptation.

In this work, we argue that this bottleneck arises from a ``brute-force'' approach to history. In many environments, transitions do not contribute equally to the learning signal; much of the interaction data is redundant, and the essential information can often be represented in a more compact form. We observe that an ideal ICRL agent should be able to maintain a fixed-size representation of the past information while remaining capable of simulating a long horizon. 
To realize this vision, we propose \textit{Recurrent Algorithm Distillation (RAD)}. As illustrated in Figure \ref{fig:RADinference}, RAD replaces the long context window of traditional AD with a dual-transformer architecture designed for recursive context compression. The system consists of: (1) a \textit{Compression Transformer} that distills long interaction sequences into a fixed-length set of latent tokens stored in the long-term memory, and (2) an \textit{AD Transformer} that generates actions by attending to a hybrid context of the latent long-term memories and recent transitions in the working memory. Crucially, as the context window fills, the system recursively recompresses its context to update the long-term memory, enabling the model to preserve information from the distant past while operating under a fixed memory budget.

We evaluate RAD across a diverse suite of RL benchmarks. Our results demonstrate that RAD generally matches or exceeds AD while using substantially shorter contexts, demonstrating that recurrent context compression provides an effective approach to scalable long-horizon ICRL.

\section{Background}

\paragraph{Partially Observable Markov Decision Processes.} 

We frame the reinforcement learning problem within a \emph{Partially Observable Markov decision process (POMDP)}, defined by the tuple $\mathcal{M} = (\mathcal{A}, \mathcal{O}, \mathcal{T}, \mathcal{R}, \gamma)$. 
At each time step $t$, the agent selects an action $a_t \in \mathcal{A}$ according to a policy $\pi$, conditioning on the current observation $o_t \in \mathcal{O}$ and the history of past interactions $ (o_0, a_0, r_0, \dots, a_{t-1}, r_{t-1})$. 
The environment receives $a_t$, transitions to a new configuration according to the transition function $\mathcal{T}$, and provides the agent with a reward $r_t$ and a subsequent observation $o_{t+1}$ based on the reward function $\mathcal{R}$ and the system's dynamics. The agent's objective is to maximize the expected return $\mathbb{E}_{\pi, M}[ \sum_{t=0}^{\infty} \gamma^t r_{t} ]$.

\paragraph{Transformers.} Transformers process sequential data using the self-attention mechanism, which allows each token to dynamically aggregate information from other tokens in the sequence. Given query, key, and value representations $Q$, $K$, and $V$, respectively, scaled dot-product attention is computed as:$$\text{Attention}(Q, K, V) = \text{softmax}\left(\frac{QK^T}{\sqrt{D}}\right)V,$$where $D$ denotes the model dimension. By directly modeling pairwise dependencies between tokens, self-attention enables Transformers to capture both local and long-range relationships without relying on recurrent computation.

\section{In-context RL and Algorithm Distillation} 

We consider the problem of \emph{in-context reinforcement learning (ICRL)}. 
In this paradigm, an agent performs gradient-free adaptation by utilizing its interaction history—the input context—to internalize environment dynamics and identify optimal behaviors. Unlike standard RL, which relies on gradient-based parameter updates, an ICRL agent leverages its context window to refine its policy during a single forward pass. 

\emph{Algorithm Distillation (AD)} \citep{laskin2022incontextreinforcementlearningalgorithm} formalizes ICRL  as a sequential prediction task. 
The core premise of AD is that the training histories of an RL algorithm inherently encode the logic of policy improvement. By modeling these histories as \emph{long history-conditioned policies} that map past experiences to subsequent actions, AD aims to induce the underlying learning rule directly from data, enabling the agent to solve novel decision-making problems without further weight updates. 

We define the \emph{context information} up to time $t$ as:
\begin{align*}
x_t = (o_0, a_0, r_0, \dots, o_{t-1}, a_{t-1}, r_{t-1}, o_t) \in \mathcal{X}\, .
\end{align*}
where $\mathcal{X}$ denotes the set of all contexts. 
An RL algorithm is formalized as a mapping $\phi: \mathcal{X} \rightarrow \Delta(\mathcal{A})$, which maps the current context information to a distribution over actions $\Delta(\mathcal{A})$. 
Given a set of training tasks $\mathcal{M}$ and a task distribution $\rho$, we sample $N$ tasks $\{M_n\}_{n=1}^N \sim \rho$ i.i.d. A dataset $\mathcal{D}$ of learning histories is then collected by running a source algorithm $\phi$ on each task for $T$ steps:
\begin{align*}
\mathcal{D} = \Big\{ (o^{n}_0, a^{n}_0, r^{n}_0, \dots,  a^{n}_{T-1}, r^{n}_{T-1}, o^{n}_T) \sim 
P_{\phi}(\cdot | M_n) \Big\}_{n=1}^N\, ,
\end{align*}
where $P_\phi( \cdot | M)$ is the probability distribution over learning sequences induced by executing algorithm $\phi$ on task $M$.  
AD optimizes a Transformer model $\pi_\theta$ by minimizing the following cross-entropy loss:
\begin{align}
{L}_{\mathrm{AD}}(\theta) = -\sum_{n=1}^N \sum_{t=0}^{T-1} \log \pi_{\theta}(a_t^n | x_{t}^n)\, .
\label{eq:ad}
\end{align}
This objective enables the transformer model $\pi_{\theta}$ to internalize the source algorithm's policy improvement patterns in-context. During deployment on a novel task, the model identifies optimal behaviors by auto-regressively sampling actions conditioned on the accumulating interaction history. 

\paragraph{AD as Posterior Sampling}
We show that AD implicitly performs {posterior sampling}.  
Define the marginal distribution over learning sequences under the task prior $\rho$: \begin{align*}
P_{\phi}(x) = \sum_{M} \rho(M) P_{\phi}(x | M) \, .
\end{align*}
The AD loss (\ref{eq:ad}) is a sampled approximation of the cross-entropy between $P_{\phi}$ and $P_{\theta}$, 
\begin{align*}
 {L}_{\mathrm{AD}}(\theta) \approx  -\sum_{M} \rho(M)  \sum_{x}  P_{\phi}(x|M)\log \pi_{\theta}(x) 
= & -\sum_{x} \sum_{M} \rho(M) P_{\phi}(x|M)\log \pi_{\theta}(x) 
\\
=& - \sum_{x} P_{\phi}(x) \log \pi_{\theta}(x) \, .
\end{align*}
Applying the chain rule for cross-entropy, we obtain:
\begin{align*}
{L}_{\mathrm{AD}}(\theta) \approx
  \sum_{t=0}^{T} \mathbb{E}_{x_t\sim P_{\phi}} \left[- \sum_{a_t} P_{\phi}(a_t | x_t ) \log \pi_{\theta}(a_t | x_t )\right] \, .
\end{align*}
The optimal model thus satisfies $\pi_{\theta^*}(a_t | x_t) = P_{\phi}(a_t | x_t)$ for all sequences in the support of $P_{\phi}$.  
By applying Bayes' rule, we can decompose this optimal policy:
\begin{align*}
 P_{\phi}(a_t | x_t) =& \frac{P_{\phi}(a_t, x_t)}{P_{\phi}(x_t)} = \frac{\sum_{M} \rho(M) P_{\phi}(a_t, x_t | M)}{\sum_{M'} \rho(M') P_{\phi}(x_t | M')} \\
 = &\sum_{M} \left( \frac{\rho(M) P_{\phi}(x_t | M)}{\sum_{M'} \rho(M') P_{\phi}(x_t | M')} \right) P_{\phi}(a_t | x_t, M) = \sum_{M} P_{\phi}(M | x_t) P_{\phi}(a_t | x_t, M)\, .
\end{align*}
This confirms that AD learns a \emph{posterior sampling policy}. 
Specifically, the model first performs implicit inference to identify the task given the current history and observation, then acts according to the source algorithm $\phi$ for that inferred task:
\begin{align}
\pi_{\theta^*}(a_t | x_t) 
=
\sum_{M} P_{\phi}(M | x_t) P_{\phi}(a_t | x_t, M)\, .
\label{eq:ad-posterior-sampling}
\end{align}

\paragraph{Scalability Bottleneck of AD} 
Practical implementations of AD typically employ a fixed-size context window, where older interactions are discarded via a sliding window mechanism once the maximum capacity is reached. 
However, this method is suboptimal for tasks involving long-range dependencies. Since ICRL relies on the context window to act as a ``{working memory}'',  the sliding window creates a truncated view of the agent's experience. 
This limitation forces the agent to rely on a localized and potentially incomplete historical subset, thereby severing the long-term temporal dependencies required for accurate posterior inference (Eq. \ref{eq:ad-posterior-sampling}). 
Consequently, this leads to inconsistent or suboptimal decision-making in complex environments where distal context is critical. 
This dependency creates a scalability bottleneck, making it difficult to extend AD to complex domains that require extensive historical context.

\section{Recurrent Algorithm Distillation}

This paper explores how to optimize the utilization of long-term historical information in algorithm distillation, addressing the inherent limitations of fixed-length context windows. 
We introduce \emph{Recurrent Algorithm Distillation (RAD)},  a novel approach featuring a dual-memory architecture. 
By maintaining a high-fidelity \emph{working memory} alongside a compressed \emph{long-term memory}, RAD allows a Transformer-based policy to condition its in-context decision-making on a significantly extended historical horizon.

\subsection{Memory Representation}

RAD maintains two complementary memory components:
\begin{itemize}[leftmargin=*]
    \item \textbf{Working memory} stores the agent's most recent transitions to capture immediate context. Let $K$ be the working memory's maximum capacity denoted by the number of transitions. At timestep $t$, let $1 \leq k \leq K$ be a lookback window, the contents of the working memory are defined as \footnote{We adopt a slight abuse of notation, letting $x_t = x_{0:t}$ represent the complete history when the lookback window $k$ spans the entire trajectory ($k=t$).}:
    \begin{align*}
x_{t-k:t} = (o_{t-k}, a_{t-k}, r_{t-k}, \dots, o_{t-1}, a_{t-1}, r_{t-1})\, .
\end{align*}

    \item \textbf{Long-term Memory} consists of latent representations designed to capture distal dependencies that extend beyond the immediate context. 
It is structured as a collection of $L$ latent tokens:
\begin{align*}
z_t = (z_{t, 0}, z_{t, 1}, \dots, z_{t, L-1})\, ,
\end{align*}
where $L$ denotes the long-term memory's maximum capacity. 
The long-term memory is initialized as an empty set, and is updated whenever the working memory reaches its maximum capacity.
\end{itemize}

The dual-memory architecture enables RAD to compress and preserve critical historical dependencies that would otherwise be lost as the working memory’s sliding window advances. 
We employ a transformer-based compression model based on Context Cascade Compression (C3) \citep{liu2025contextcascadecompressionexploring}. 
Given an input sequence of tokens $(y_1, \dots, y_m)$, the compressor $q_{\omega}$ produces a compressed sequence of tokens $ (y'_1, \dots, y'_{m'})$ conditioned on $m'$ learnable query tokens with $m' \ll m$. This compression is achieved by jointly training a reconstruction model-also a Transformer-which aims to recover the original sequence $(y_1, \ldots, y_m)$ conditioned on the latent tokens $(y_1', \ldots, y_{m'}')$

\subsection{Tokenization}
\label{tokenization}

In contrast to \citet{lee2023supervisedpretraining} and \citet{son2025distillingreinforcementlearningalgorithms}'s tokenizer which embeds each transition $(o_{t-1}, a_{t-1}, r_{t-1}, o_t)$ into a single embedding vector, at timestep $t$, we treat observation $o_t$, action $a_t$, reward $r_t$ as distinct tokens, and each is embedded separately into $\mathbb{R}^{d}$. Each timestep therefore contributes three tokens. To ensure architectural compatibility, the long-term memory is likewise composed of $d$-dimensional latent tokens, allowing the latent state to be directly concatenated with tokenized transitions, forming a unified input sequence for both the policy $\pi_\theta$ and the compression model $q_w$. To preserve the temporal structure, we set $L$ to a multiple of 3, ensuring that compression does not split the $(o_t, a_t, r_t)$ tuple at any timestep $t$.

\subsection{Inference}
\label{sec:inference}

RAD autoregressively generates actions conditioned on both memory components. At timestep $t$, given the current long-term memory $z$, the action is sampled as: 
\[
a_t \sim \pi_{\theta}(\cdot\mid z,x_{t-k:t}, o_t).
\]

RAD then takes an environment step, receiving reward $r_t$ and the next observation $o_{t+1}$. $a_t$ and $r_t$ are appended to the working memory, after which RAD proceeds to predict the next action.

When the working memory reaches its capacity $k=K$, we perform a \textit{compression} update to \textit{distill} critical information from the current context into a new latent state. A compression Transformer $q_\omega$ produces a candidate latent state $c$ from the current context:
\[
c = q_\omega(z, x_{t-K:t})
\in\mathbb{R}^{L\times d}.
\]
The candidate is integrated with the previous long-term memory using a GRU-style gated update. Denote $v = [z, c]$ as the concatenation along the feature dimension at each latent position. The new latent state is updated as:
\[
z \leftarrow (1-g) \odot z + g \odot (c + \delta),
\]
where $g=\sigma(W_gv + b_g)$ and $\delta=\tanh(W_\delta v + b_\delta)$. Particularly, at the first compression event, since the long-term memory is empty, we bypass the gated update and directly set $z \leftarrow q_\omega(x_{0:K})$.

Immediately following the latent state update, the lookback window is set to $p$ where $p << K$, preserving a small window of recent context across memory updates while reserving context for subsequent interactions. This inference procedure is detailed in Algorithm \ref{alg:rad-inference}.

\subsection{Training}

RAD utilizes the same dataset $\mathcal{D}$ as standard Algorithm
Distillation (AD), constructed by collecting the source RL
algorithm's learning histories across multiple tasks. 
We jointly optimize the policy, compressor, and gated memory update through action
prediction. Consider a sampled training sequence $x_{t_0:t_0+\ell}$,
where $t_0$ is its starting timestep and
$\ell=K+nD$. Here, $K$ is the working-memory capacity, $p<K$
is the number of transitions retained after compression, $n$
is the number of compression events, and $D=K-p$ is the number of transitions advanced per compression.

The history incorporated into long-term memory is
partitioned into $(n+1)$ contiguous segments:
\[
x_{t_0:t_0+D},\ x_{t_0+D:t_0+2D},\ \dots,\
x_{t_0+(n-1)D:t_0+nD}, x_{t_0+nD:t_0+\ell}
\]

We recursively apply the compressor
$q_{\omega}$ and gated update $U_{\phi}$ defined in
Section~\ref{sec:inference}. 
Starting from empty memory $z_0=\varnothing$, we obtain latent states
$(z_1,\ldots,z_n)$ by
\begin{align}
z_1 &= q_\omega(x_{t_0, t_0+K}), \\
z_i &= U_{\phi}(z_{i-1},q_\omega(z_{i-1}, x_{t_0+(i-1)D : t_0+(i-1)D+K})), \qquad i=2,\ldots,n
\end{align}

The policy is then trained to predict the source learner's actions
throughout the final working memory $x_{t_0+nD:t_0+\ell}$. At timestep
$t$, it conditions on the final latent state $z_n$, the preceding
working-memory history $x_{t_0+nD:t}$, and the current observation
$o_t$. The RAD objective is
\begin{equation}
\mathcal{L}_{\mathrm{RAD}}(\theta,\omega,\phi)
=
-\mathbb{E}_{x_{t_0:t_0+\ell}\sim\mathcal{D}}
\left[
\frac{1}{K}
\sum_{t=t_0+nD}^{t_0+\ell-1}
\log \pi_{\theta}
\left(a_t \mid z_n,x_{t_0+nD:t},o_t\right)
\right],
\end{equation}
where the expectation is over training sequences sampled from
the learning histories in $\mathcal{D}$, including their starting
timesteps and number of compression events.

To bound the cost of backpropagation through the recurrent memory,
we retain gradients through only the most recent $G$ compression
events.
The training procedure is detailed in
Algorithm~\ref{alg:rad-training}.

\begin{figure}
    \centering
    \includegraphics[width=1.0\linewidth]{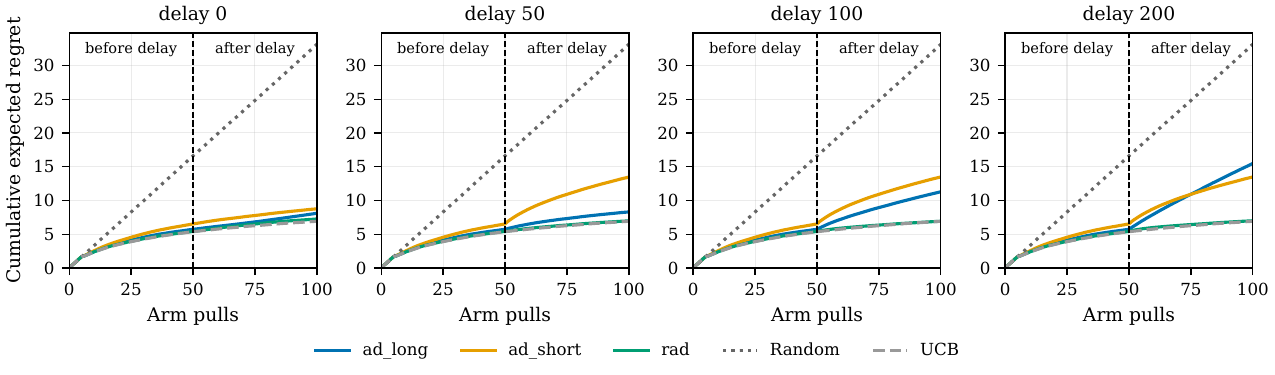}
    \caption{Cumulative regret on Delayed-Bandit. Panels show delays of \(D\in\{0,50,100,200\}\) zero-reward distractor steps. Results are averaged over 5 training seeds and 5 evaluation seeds.}
    \label{fig:bandit-cumulative-regret}
\end{figure}

\section{Experiments}

\subsection{Environments}

We evaluate the performance of RAD across a diverse set of discrete and continuous environments characterized by varying task complexities and horizon lengths.

\paragraph{Delayed-Bandit} Delayed-Bandit evaluates the ability to retain useful information across interruptions in a stationary multi-armed bandit task. The base bandit setting follows \cite{lee2023supervisedpretraining}'s design, where each arm has a fixed mean sampled independently from $\mathcal{U}[0,1]$, with rewards drawn from $\mathcal{N}(\mu_a, 0.3^2)$, and a bandit algorithm runs for 100 arm pulls. However, Delayed-Bandit inserts distractions into the learning histories. The resulting trajectories each consists of 50 arm pulls, followed by $N_\text{delay}$ distractor transitions with uniformly sampled actions and zero rewards, and then 50 additional pulls on the same task.

\paragraph{Grid-World} We consider two discrete grid-world maze environments: \textbf{Darkroom} and \textbf{Dark Key-to-Door}. In Darkroom, the agent starts at the center of a $9 \times 9$ grid and must locate a hidden goal while observing only its current coordinates. The 81 tasks, defined by distinct goal locations, are partitioned into a 9:1 train-test split. Each episode lasts 20 steps, and the agent receives a reward of 1 upon reaching the goal. Dark Key-to-Door extends this setting by requiring the agent to find a key before reaching the goal, yielding 6561 distinct task configurations and an episode horizon of 50 steps. The agent receives a reward of 1 for finding the key and an additional reward of 1 for reaching the goal, for a maximum episode reward of 2.

\paragraph{Meta-World} The Meta-World robotic manipulation benchmark suite \cite{mclean2025metaworld} is employed to evaluate RAD's performance. We utilize the ML1 tasks, which focus on meta-learning a single task type across 50 different seeds representing varied object and goal locations. These tasks feature horizons of 100 steps.

\subsection{Dataset Generation}

Similar to AD, we first generated a dataset for each environment which consists of training histories of the source RL algorithm solving the training tasks. In this research, we used UCB \citep{lai1985asymptotically} for bandits and PPO \citep{schulman2017proximalpolicyoptimizationalgorithms} for MDPs. Particularly, for Dark Key-to-Door, due to difficulty of exploration in sparse reward settings, we stacked the past 8 observations for the PPO agent as input, but saving only the true observation at the current timestep to ensure compatibility in the training dataset.

\begin{figure}[t]
    \centering
    \includegraphics[width=0.8\linewidth]{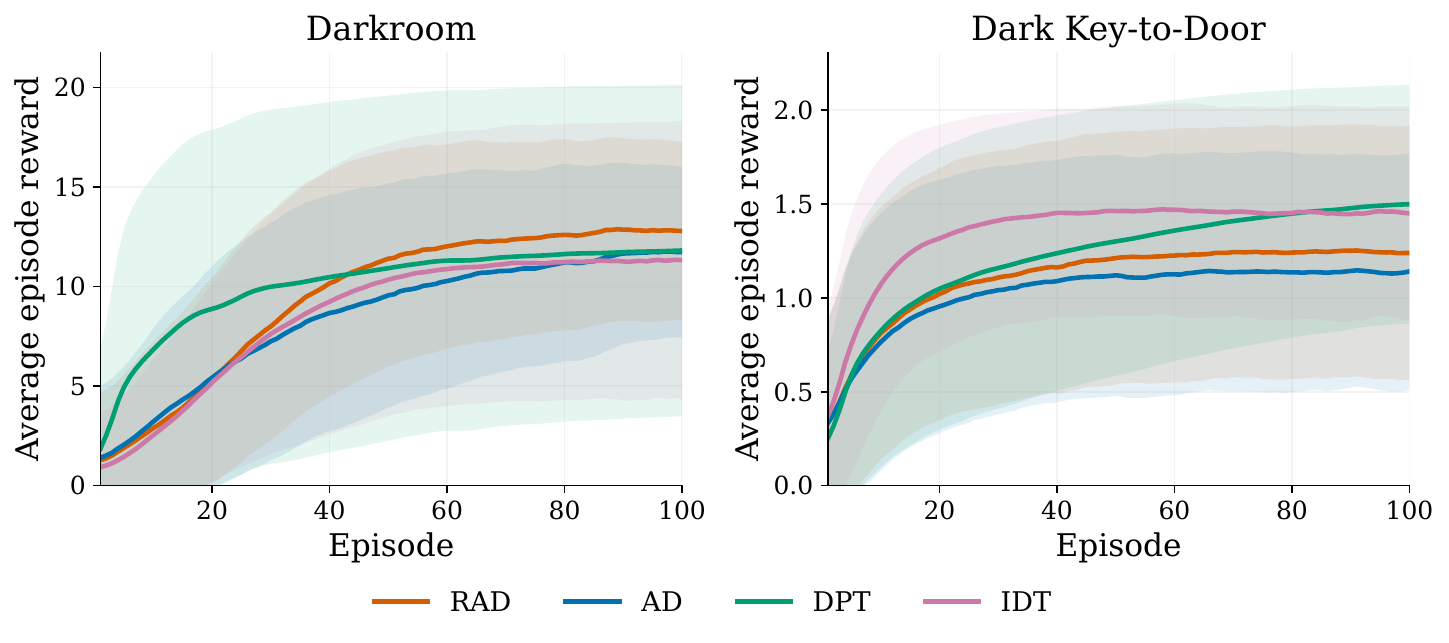}
    \caption{Average episode reward in Darkroom and Dark Key-to-Door, comparing RAD, AD, DPT \citep{lee2023supervisedpretraining}, and IDT \citep{huang2024context}. Results are averaged over 3 training seeds and 20 evaluation seeds. RAD outperforms all baseline models in Darkroom, and also gains higher rewards than AD in Dark Key-to-Door. DPT and IDT achieve higher rewards in Dark Key-to-Door, but both require domain knowledge including optimal actions and desired returns respectively, which generally are not acquirable in regular tasks. On the contrary, though constrained by the high sub-optimality of the source RL algorithm, RAD and AD can autonomously generalize without additional information.}
    \label{fig:gridworld}
\end{figure}

\subsection{Results}

The main research question is whether AD can incorporate with a long-term memory buffer and reinforcement learn in context but with a shorter context window. The following are analyzed.

\paragraph{Can RAD effectively perform in-context reinforcement learning on tasks where keeping history is important?}
We first collect source data for Delayed-Bandit by running UCB \citep{lai1985asymptotically} for 1000 train tasks, insert $N_\text{delay} = 50$ distractor transitions for each trajectory, and then train AD and RAD. We compare two AD variants: $\mathrm{AD}_{\text{short}}$ with the same context window as RAD, and $\mathrm{AD}_{\text{long}}$ with a window exceeding the training trajectory length. We evaluate all models on 100 held-out tasks across varying delay lengths.

Figure \ref{fig:bandit-cumulative-regret} shows that all models effectively distill UCB for standard multi-armed bandit tasks with no delay. However, once delay takes place, $\text{AD}_\text{short}$ loses pre-delay experience from its sliding window and is forced to re-explore. Both $\mathrm{AD}_{\text{long}}$ and RAD retain useful history through full-context attention and latent memory respectively when $N_\text{delay}=50$, with RAD achieving lower regret. Additionally, RAD generalizes across unseen delays, consistently matching UCB for $N_\text{delay}=100$ and $200$, whereas $\mathrm{AD}_{\text{long}}$ degrades substantially. These results suggest that RAD effectively learns latent representations for long-term memories, which is able to filter noise and preserve essential information for action prediction.

\paragraph{Can RAD explore, assign credit, and generalize in MDP settings?}
As shown in Figure \ref{fig:gridworld}, RAD outperforms AD on Grid-World tasks despite a potentially shorter active context. Like AD, RAD autonomously generalizes to unseen Grid-World tasks, learning in context to explore, assign credit from sparse rewards, and exploit acquired knowledge without parameter updates. These results suggest that RAD's recursive compression preserves the ``improvement operator'' learned through distillation.

\begin{figure}[t]
    \centering
    \includegraphics[width=1.0\linewidth]{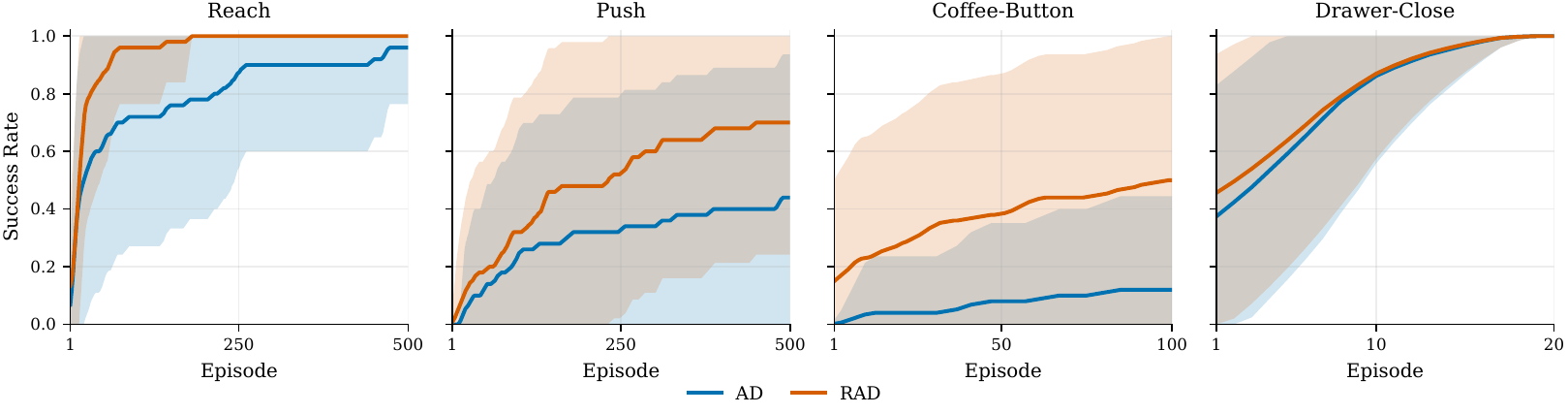}
    \caption{Success rates on Meta-World tasks. Each environment is trained on 50 training tasks, and results are averaged over 50 testing tasks.}
    \label{fig:metaworld}
\end{figure}

\paragraph{Does RAD solve continuous-controll tasks effectively?} 
Meta-World features a set of continuous robot manipulation tasks with large state and action spaces, emphasizing on generalization across observations and actions. Shown in Figure \ref{fig:metaworld}, RAD generally matches and outperforms AD on the tasks like Reach and Push, demonstrating its ability to maintain a useful long-term memory in high-dimensional continuous settings.

\paragraph{How computationally efficient is RAD?}
We profile inference across all experiments and report average FLOPs per action prediction in Table~\ref{tab:inference_flops}. RAD reduced computation for more than 80\% in most environment (except for Dark Key-to-Door due to the extreme sparsity in rewards and the hardness of preserving useful information), indicating that RAD provides a scalable solution for efficient in-context decision-making. 

\begin{table}[h]
\centering
\caption{Inference FLOPs per action for each sequence.}
\label{tab:inference_flops}
\begin{tabular}{lrrr}
\toprule
\textbf{Environment} & \textbf{AD FLOPs / Action} & \textbf{RAD FLOPs / Action} & \textbf{Compute Ratio} $\downarrow$ \\
\midrule
Bandit               & 59.8 M (short)  & 43.0 M  & \textbf{0.72$\times$} \\
 & 263.4 M (long) & & \textbf{0.16$\times$}\\
Darkroom & 144.8 M & 27.8 M  & \textbf{0.19$\times$} \\
Dark Key-to-Door & 203.4 M & 85.9 M & \textbf{0.42$\times$} \\
Meta-World     & 1.84 G  & 130.6 M & \textbf{0.07$\times$} \\
\bottomrule
\end{tabular}
\end{table}

\section{Ablation Studies}

We evaluate and discuss the key design choices in RAD. More details are discussed in Appendix \ref{appendix:ablation}.


\noindent
\begin{minipage}[t]{0.56\textwidth}
\vspace{0pt}

\paragraph{Effect of the Long-term Memory Update Gate.}

We compare the following long-term memory update methods on Darkroom:

\begin{itemize}[leftmargin=1.5em]
    \item \textbf{Replace:} Directly adopt the candidate:
    \(z \leftarrow c\).

    \item \textbf{Residual:} Add the candidate to the old latent state
    and normalize:
    \(z \leftarrow \operatorname{LayerNorm}(z+c)\).

    \item \textbf{Multiplicative:} Compute a gate from old latents to
    scale each candidate feature:
    \(z \leftarrow \sigma(W_gz+b_g) \odot c\).

    \item \textbf{GRU:} The method adopted in this work.
\end{itemize}

Figure~\ref{fig:latent_update_ablation} shows that GRU gating yields a higher median reward and a narrower interquartile range, supporting its use for recurrent memory updates, possibly due to its role in stabilizing gradient flow during back-propagation through the recurrent memory.

\paragraph{Comparison of Different Tokenizers.}

We compare our tokenization method with \cite{lee2023supervisedpretraining}'s 
DPT-style tokenizer as introduced in Section \ref{tokenization}. As illustrated in Figure \ref{fig:ablation_tokenization}, both tokenization schemes achieved comparable returns for AD and RAD. 

However, this work's tokenizer supports more efficient training since its context grows by appending tokens between compression steps, allowing a single Transformer forward pass to supervise multiple observation positions. On the contrary, DPT-style tokenization reconstructs the context at each timestep during inference and therefore can only supervise one observation per input sequence during training.

\end{minipage}
\hfill
\begin{minipage}[t]{0.40\textwidth}
\vspace{0pt}
\centering

\includegraphics[
    width=\linewidth
]{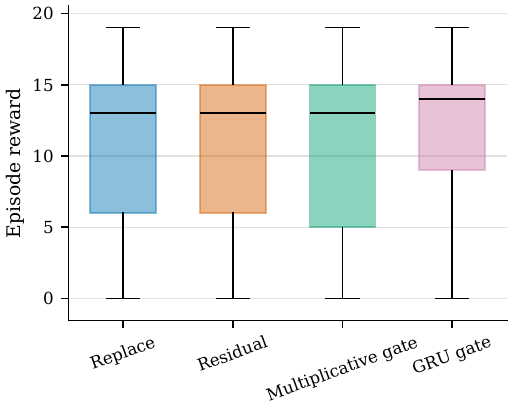}

\captionof{figure}{
    Comparison of different long-term memory update methods.
}
\label{fig:latent_update_ablation}

\vspace{4pt}
\centering

\includegraphics[
    width=\linewidth
]{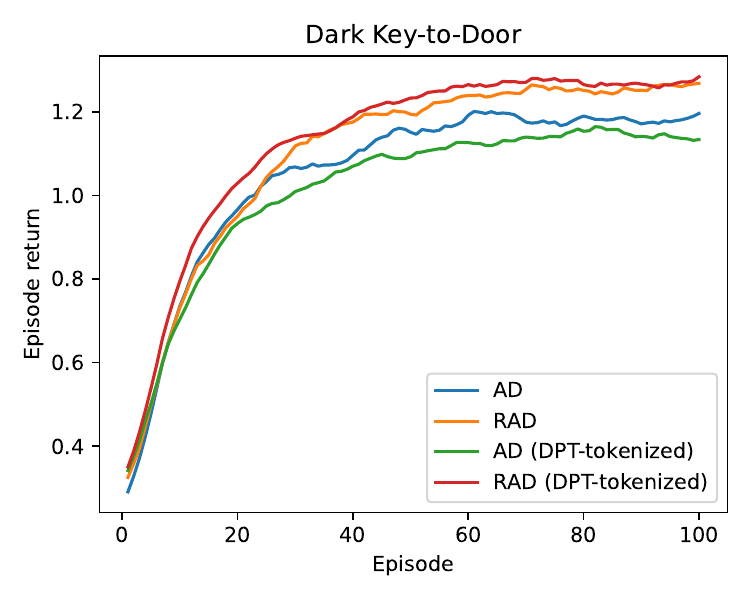}

\captionof{figure}{
    Performance of AD/RAD with different tokenizers on Dark Key-to-Door.
}
\label{fig:ablation_tokenization}

\end{minipage}

\section{Related Work}

\paragraph{In-Context Reinforcement Learning.}
Transformers have been widely applied to offline RL and sequential
decision-making
\citep{chen2021decision,janner2021offline,
furuta2022generalizeddecisiontransformeroffline,
lee2022multigame,reed2022generalistagent}.
Beyond learning a fixed policy, in-context RL (ICRL) aims to adapt to new
tasks from interaction history without parameter updates.
Algorithm Distillation (AD)
\citep{laskin2022incontextreinforcementlearningalgorithm}
trains a Transformer on multi-episode RL learning histories to reproduce its learning process
in context.
Decision-Pretrained Transformer (DPT)
\citep{lee2023supervisedpretraining} instead learns to infer optimal
actions directly from interaction data.
Subsequent work has improved ICRL by injecting noise into training data
\citep{zisman2024emergenceincontextreinforcementlearning},
supporting variable action spaces
\citep{sinii2024incontextreinforcementlearningvariable},
scaling up high-quality training tasks
\citep{wang2026towards},
introducing mixture-of-experts architectures
\citep{wu2026mixture},
reweighting action-prediction losses
\citep{dong2026context},
and adopting richer prediction objectives, including transition and
action-value functions
\citep{son2025distillingreinforcementlearningalgorithms,
Mukherjee_Pretraining_2025,liu2026scalable}.
These works primarily improve how in-context policies are learned, whereas
RAD focuses on extending the interaction history that can be retained under
a fixed memory budget.

\paragraph{Long-Context Modeling and Memory.}
Handling long sequences efficiently has been a longstanding challenge for
Transformers.
One line of work reduces the computational cost of directly processing long
contexts through sparse or linear attention
\citep{beltagy2020longformerlongdocumenttransformer,
zaheer2021bigbirdtransformerslonger,
katharopoulos2020transformersrnnsfastautoregressive,
yang2025gated,team2025kimi}.
Another line instead propagates information across bounded segments.
Transformer-XL
\citep{dai2019transformerxlattentivelanguagemodels}
caches hidden states from previous segments,
while Compressive Transformer
\citep{rae2019compressivetransformerslongrangesequence}
further compresses evicted activations into a lower-resolution memory. Block-Recurrent Transformer \citep{hutchins2022block} maintains recurrent state vectors with gated updates across blocks, and Recurrent Memory Transformer
\citep{bulatov2022recurrentmemorytransformer}
takes a different approach by propagating learned memory tokens between
segments, maintaining a fixed-size recurrent representation of past context. More recent approaches such as Infini-attention \citep{munkhdalai2024leave} combine local attention with bounded compressive memory, while Titans \citep{behrouz2026titans} combines attention-based short-term context with persistent neural long-term memory. Retrieval-based approaches such as Memorizing Transformers \citep{wu2022memorizing} instead maintain an external store of past representations.

Long-context mechanisms have subsequently been applied to reinforcement
learning, where agents must reason over extended interaction histories.
GTrXL
\citep{parisotto2019stabilizingtransformersreinforcementlearning}
adapts Transformer-XL-style recurrence to RL,
while RATE \citep{cherepanov2026recurrent} and ELMUR \citep{cherepanov2026elmur}
augment Transformer policies with recurrent or structured external memories for long-horizon decision making. Elastic Decision Transformer (EDT) \citep{wu2023elastic} dynamically adjusts the amount of provided trajectory history to facilitate trajectory stitching.
For ICRL, Lu et al.~\citep{lu2023structured} employ structured state-space models to summarize interaction history in a recurrent latent state, while AMAGO \citep{grigsby2024amago}, in contrast, retains a Transformer-based agent and enables end-to-end off-policy RL over long interaction sequences, relying on an explicitly long context as the agent's memory. RA-DT \citep{schmied2024retrieval} addresses the context bottleneck through retrieval, storing past experiences in an external memory and selecting relevant sub-trajectories for the current decision.
RAD takes a complementary approach: rather than retaining or retrieving an expanding set of past experiences, it preserves recent interactions at full fidelity while recursively consolidating older learning history into a fixed-size latent state optimized for in-context action prediction.

\section{Conclusion}

This study has illustrated that Recurrent Algorithm Distillation provides a scalable solution for In-Context Reinforcement Learning. Through a dual-memory architecture with recursive compression, RAD enables efficient long-horizon inference while reducing memory and computational requirements. A key limitation is training stability, as recurrent memory updates require backpropagation through time, increasing optimization and computation difficulty over long horizons. Future work could investigate more stable training objectives, alternative memory-update mechanisms, or approaches that reduce the need for long-range gradient propagation. Overall, RAD provides a promising direction toward long-horizon ICRL agents that retain useful historical information under bounded computational and memory budgets.

\bibliography{iclr2027_conference}
\bibliographystyle{iclr2027_conference}

\newpage
\appendix
\section{RAD Algorithm Details}
\label{appendix:algorithmdetail}
The following are the pseudocodes for RAD, including inference and training.

\begin{algorithm}[h]
\caption{RAD Inference}
\label{alg:rad-inference}
\begin{algorithmic}[1]
\Require AD Transformer $\pi_\theta$, Compression Transformer $q_\omega$,
         Update Gate $U_\phi$, $K$, $p$
\State $i \gets 0$, $t \gets 0$, $x \gets ()$, $z_0 \gets \varnothing$

\While{not terminal}
    \State $a_t \sim \pi_\theta(\cdot \mid z_i, x, o_t)$
    \State $(r_t, o_{t+1}) \gets \Call{Environment}{a_t}$
    \State $x \gets x \mathbin{\|} (o_t, a_t, r_t)$

    \If{$|x| = K$}
        \If{$i = 0$}
            \State $z_{i+1} \gets q_\omega(x)$
        \Else
            \State $z_{i+1}
                \gets U_\phi\!\left(
                    z_i,\,
                    q_\omega(z_i, x)
                \right)$
        \EndIf
        \State $x \gets x_{K-p+1:K}$
        \State $i \gets i + 1$
    \EndIf

    \State $t \gets t + 1$
\EndWhile
\end{algorithmic}
\end{algorithm}

\begin{algorithm}[h]
\caption{RAD Training}
\label{alg:rad-training}
\begin{algorithmic}[1]
\Require Training dataset $\mathcal{D}$, AD Transformer $\pi_\theta$,
         Compression Transformer $q_\omega$, Update Gate $U_\phi$,
         Reconstruction Model $d_\psi$, $K$, $p$
\State $D \gets K-p$

\While{not converged} \Comment{\textbf{Phase 1: Compression Pretraining}}
    \State Sample $x_{t-k:t} \sim \mathcal{D}$
    \State Update $(\omega,\psi)$ by minimizing
    \[
        \mathcal{L}_{\mathrm{pre}}
        =
        \mathbb{E}_{x_{t-k:t} \sim \mathcal{D}}
        \left[
            \left\|
                d_\psi\!\left(q_\omega(x_{t-k:t})\right)-y
            \right\|_F^2
        \right]
    \]
\EndWhile

\While{not converged} \Comment{\textbf{Phase 2: Action Distillation}}
    \State Sample $x_{t_0:t_0+\ell} \sim \mathcal{D}$,
           where $\ell = K+nD$

    \For{$i = 1,\ldots,n$}
        \If{$i = 1$}
            \State $z_i \gets q_\omega(x_{t_0:t_0+K})$
        \Else
            \State $z_i \gets
                U_\phi\!\left(
                    z_{i-1},
                    q_\omega\!\left(
                        z_{i-1},
                        x_{t_0+(i-1)D:t_0+(i-1)D+K}
                    \right)
                \right)$
        \EndIf
    \EndFor

    \State Compute
    \[
        \mathcal{L}_{\mathrm{RAD}}
        \gets
        -\frac{1}{K}
        \sum_{t=t_0+nD}^{t_0+\ell-1}
        \log
        \pi_\theta
        \left(
            a_t
            \mid
            z_n,\,
            x_{t_0+nD:t},\,
            o_t
        \right)
    \]

    \State Update $(\theta,\omega,\phi)$ by minimizing
           $\mathcal{L}_{\mathrm{RAD}}$
\EndWhile
\end{algorithmic}
\end{algorithm}

\section{Experiment Details}
\label{app:experiment-details}

This appendix specifies the architectures, source RL
algorithms, and training configurations used for training RAD. We distinguish an environment
transition from a Transformer token: each completed transition contributes three tokens, corresponding to its observation, action, and reward. A latent memory token is a learned vector and does not correspond to an individual transition in token length.

\subsection{Model Architectures}
\label{app:model-architectures}

\paragraph{AD Transformer.}
The policy backbone is a GPT-2-style decoder-only Transformer that predicts actions from
the preceding interaction history. We embed observations, actions, and rewards
separately and interleave their embeddings in temporal order,
$(o_0,a_0,r_0,o_1,a_1,r_1,\ldots,o_t)$. Learned token-type embeddings distinguish
the three modalities, and learned absolute positional embeddings encode their
positions within the current context. Grid-World observations are embedded
using a lookup table over grid positions. Delayed-Bandit uses an embedding
table for the two observation types, genuine interaction and distraction.
For Meta-World, a linear layer embeds the 11-dimensional observation used by
the sequence model. Discrete actions are one-hot encoded and linearly
projected; continuous actions and scalar rewards have separate linear
projections.

Each Transformer block consists of multi-head self-attention and a two-layer
feed-forward network with GELU activation, residual connections, and
pre-layer normalization. A final layer normalization precedes a linear
action-prediction head. Action predictions are read from observation-token
outputs, with causal attention over the uncompressed history. 

\paragraph{Compression Transformer.}
RAD augments the policy backbone with a query-based Compression Transformer.
Given embedded history tokens, $m$ learned query tokens are appended to the sequence and the sequence gets forwarded into the Compression Transformer.
Each compression layer
applies self-attention among the queries, cross-attention from the queries to the input history, and a GELU feed-forward network. Each sublayer uses pre-layer normalization and a residual connection. A final layer normalization is applied to output the candidate updates.

\subsection{Source Reinforcement-Learning Algorithms}
\label{app:source-algorithms}

We construct the offline datasets from the interaction histories produced
while the source algorithms learn each task. These histories preserve the
temporal order of experience and the progression of the source learner.

\paragraph{Upper Confidence Bound (UCB).}
UCB balances exploitation of arms with high empirical
rewards and exploration of less frequently sampled arms.
We use the UCB variant with the exploration bonus used in MBIE-EB \citep{strehl2008analysis}. After pulling
each arm once in index order, the algorithm selects
\begin{equation}
 a_t=\mathop{\arg\max}_{a}\left[
 \widehat\mu_a+\frac{\beta}{\sqrt{N_a}}\right],\qquad \beta=1,
\end{equation}
where $N_a$ counts genuine pulls of arm $a$ and $\widehat\mu_a$ is its
empirical mean reward. Ties are broken in favor of the lowest arm index.
During the distractor interval, actions are sampled uniformly and rewards
are zero. These transitions remain in the model's input history, while the
UCB statistics remain unchanged. Only genuine arm pulls contribute action
targets during distillation.

\paragraph{Proximal Policy Optimization (PPO).}
For Grid-World and Meta-World, we use the Stable-Baselines3 \citep{raffin2021stable-baselines3} implementation
of PPO \citep{schulman2017proximalpolicyoptimizationalgorithms}. PPO is an on-policy actor--critic method that
alternates between collecting rollouts and optimizing a clipped policy
surrogate. With probability ratio
$\rho_t(\theta)=\pi_\theta(a_t\mid o_t)/\pi_{\theta_{\mathrm{old}}}(a_t\mid o_t)$,
its policy objective is
\begin{equation}
 \mathcal{J}_{\mathrm{clip}}(\theta)=
 \mathbb{E}_t\!\left[
 \min\!\left(\rho_t(\theta)\widehat A_t,
 \operatorname{clip}(\rho_t(\theta),1-\epsilon,1+\epsilon)\widehat A_t\right)
 \right].
\end{equation}
Advantages are estimated using Generalized Advantage Estimation (GAE) \citep{schulman2015high} and
normalized before policy updates. The value function is trained by
regression to the estimated returns. The actor and critic each use two
hidden layers of 64 units with tanh activations. We use a categorical policy
for Grid-World and a diagonal-Gaussian policy for Meta-World. Each source
learner collects experience from 100 parallel environment streams for its
task. The histories of these streams are retained for sequence-model
training.

\subsection{Training Details}
\label{app:training-details}

\paragraph{Hardware and optimization.}
Experiments were run using eight NVIDIA GeForce RTX 4090 GPUs. Sequence
models are implemented in PyTorch. We optimize the sequence models with AdamW \citep{loshchilov2017decoupled}, using $(\beta_1,\beta_2)=(0.9,0.99)$, weight decay $0.01$, and gradient clipping
at an $\ell_2$ norm of $1.0$. Training uses
linear learning-rate warmup followed by cosine decay. Table~\ref{tab:training-config}
reports the configured batch sizes and optimization budgets.

\paragraph{Curriculum.}
For Grid-World and Meta-World, policy training gradually increases the
allowed number of compression events and shifts sampling toward longer
histories. Each training batch uses a single compression-count bucket.
Let $c$ denote the number of compressions. We group counts into short ($c=0$), medium
($c=1$--$2$), long ($c=3$--$8$), and very long ($c\geq9$) categories.
Within a category, probability is divided equally among the available
counts; probability assigned to an unavailable harder category is added
to the largest allowed count. Table~\ref{tab:curriculum} gives the stage
boundaries, compression limits, and category probabilities.

At each curriculum transition, the learning rate warms up from 10\% of
its group-specific peak to that peak, then decays cosinusoidally toward
10\% over the remainder of the stage. Initial warmup lasts 2,000 steps;
later stage warmups last 1,000 steps for Darkroom and Meta-World and
2,000 steps for Dark Key-to-Door. Delayed-Bandit uses no staged compression
curriculum and samples genuine-action targets uniformly over the dataset.

\subsection{Hyperparameters}
\label{app:hyperparameters}

Tables~\ref{tab:ad-rad-hyperparameters}--\ref{tab:curriculum} provide the
architecture, context, source-algorithm, and training settings. DR denotes Darkroom and DKTD denotes Dark Key-to-Door. 

\begin{table}[htbp]
\centering
\small
\setlength{\tabcolsep}{5pt}
\caption{AD and RAD hyperparameters. RAD uses the same policy architecture as AD. Context windows and capacities are measured in tokens unless otherwise specified (1 transition = 3 tokens).}
\label{tab:ad-rad-hyperparameters}
\begin{tabular}{lrrrr}
\toprule
Hyperparameter & Delayed-Bandit & DR & DKTD & Meta-World \\
\midrule
\multicolumn{5}{l}{\textbf{AD baseline}} \\
Policy model dimension & 64 & 64 & 64 & 64 \\
Policy Transformer layers & 4 & 4 & 4 & 4 \\
Policy attention heads & 4 & 4 & 4 & 8 \\
Policy feed-forward dimension & 256 & 256 & 256 & 256 \\
Attention / residual dropout & 0.1 & 0.1 & 0.1 & 0.1 \\
Policy context window & 150 / 900 & 240 & 300 & 1200 \\
\midrule
\multicolumn{5}{l}{\textbf{RAD}} \\
Compression model dimension & 64 & 64 & 64 & 64 \\
Compression Transformer layers & 2 & 3 & 4 & 4 \\
Compression attention heads & 4 & 4 & 4 & 4 \\
Compression feed-forward dimension & 256 & 256 & 256 & 256 \\
Compression dropout & 0.1 & 0.1 & 0.1 & 0.1 \\
Latent memory tokens $m$ & 15 & 15 & 60 & 60 \\
Working memory size (transition) $K$ & 50 & 25 & 52 & 80 \\
Retained recent transitions $p$ & 5 & 5 & 10 & 20 \\
Gradient-bearing compression rounds $G$ & $\infty$ & 5 & 5 & 6 \\
Policy context window & 165 & 90 & 216 & 300 \\
\bottomrule
\end{tabular}
\end{table}

\begin{table}
\centering
\small
\setlength{\tabcolsep}{6pt}
\caption{PPO source-algorithm configuration.}
\label{tab:ppo-config}
\begin{tabular}{lrrr}
\toprule
Hyperparameter & DR & DKTD & Meta-World \\
\midrule
Parallel streams per learner & 100 & 100 & 100 \\
Rollout steps per stream & 80 & 50 & 100 \\
Optimization minibatch size & 40 & 100 & 200 \\
Epochs per rollout update & 20 & 10 & 20 \\
Total Timesteps & 100,000 & 100,000 & 1,000,000 \\
Learning rate & $3\!\times\!10^{-4}$ & $3\!\times\!10^{-4}$ & $3\!\times\!10^{-4}$ \\
Discount factor $\gamma$ & 0.99 & 0.99 & 0.99 \\
GAE parameter $\lambda$ & 0.95 & 0.95 & 0.95 \\
Policy clip range $\epsilon$ & 0.2 & 0.2 & 0.2 \\
Value-loss coefficient & 0.5 & 0.5 & 0.5 \\
Entropy coefficient & 0 & 0 & 0 \\
Gradient clip norm & 0.5 & 0.5 & 0.5 \\
\bottomrule
\end{tabular}
\par\vspace{3pt}
\begin{minipage}{0.98\linewidth}\footnotesize
All three configurations use tanh activations, Adam with its default
$(\beta_1,\beta_2)=(0.9,0.999)$, optimizer epsilon $10^{-5}$, and no weight
decay. Implementations are inherited from Stable-Baselines3 \citep{raffin2021stable-baselines3} and PyTorch \citep{paszke2019pytorch}.
\end{minipage}
\end{table}

\begin{table}[htbp]
\centering
\small
\setlength{\tabcolsep}{5pt}
\caption{AD/RAD training hyperparameters}
\label{tab:training-config}
\begin{tabular}{lrrrr}
\toprule
Hyperparameter & Delayed-Bandit & DR & DKTD & Meta-World \\
\midrule
\multicolumn{5}{l}{\textit{AD policy training}} \\
Training steps & 50k / 100k$^a$ & 50k & 50k & 50k \\
Configured batch size & 64 & 512 & 512 & 256 \\
Peak learning rate & $3\!\times\!10^{-4}$ & $3\!\times\!10^{-4}$ & $3\!\times\!10^{-4}$ & $3\!\times\!10^{-4}$ \\
Warmup steps & 500 & 1,000 & 1,000 & 5,000 \\
\midrule
\multicolumn{5}{l}{\textit{RAD compression pretraining}} \\
Training steps & 20k & 40k & 50k & 50k \\
Configured batch size & 64 & 512 & 1,024 & 256 \\
Peak learning rate & $10^{-4}$ & $3\!\times\!10^{-4}$ & $3\!\times\!10^{-4}$ & $3\!\times\!10^{-4}$ \\
Warmup steps & 500 & 3,000 & 2,000 & 3,000 \\
\midrule
\multicolumn{5}{l}{\textit{RAD policy training}} \\
Training steps & 100k & 100k & 100k & 100k \\
Configured batch size & 64 & 256 & 1,024 & 128 \\
Policy / compressor / latent LR$^b$ & 3 / 1 / 3 & 3 / 1 / 3 & 3 / 1 / 3 & 3 / 1 / 3 \\
Initial warmup steps & 500 & 2,000 & 2,000 & 2,000 \\
Later stage warmup steps & --- & 1,000 & 2,000 & 1,000 \\
Learning-rate floor / peak & 0 & 0.1 & 0.1 & 0.1 \\
\bottomrule
\end{tabular}
\par\vspace{3pt}
\begin{minipage}{0.98\linewidth}\footnotesize
$^a$The two values correspond to AD-Short / AD-Long. \\
$^b$All three learning rates are in units of $10^{-4}$.

\end{minipage}
\end{table}

\begin{table}[htbp]
\centering
\small
\setlength{\tabcolsep}{5pt}
\caption{RAD curriculum. Probabilities correspond to short, medium, long,
and very-long compression-count categories.}
\label{tab:curriculum}
\begin{tabular}{llrrrrrr}
\toprule
Environment & Start step & Max. $c$ & Short & Medium & Long & Very long \\
\midrule
DR & 0 & 1 & 0.60 & 0.35 & 0.05 & 0.00 \\
DR & 30k & 3 & 0.35 & 0.40 & 0.20 & 0.05 \\
DR & 50k & 6 & 0.25 & 0.30 & 0.30 & 0.15 \\
DR & 75k & $\infty$ & 0.25 & 0.25 & 0.25 & 0.25  \\
\midrule
DKTD & 0 & 2 & 0.30 & 0.55 & 0.15 & 0.00  \\
DKTD & 25k & 4 & 0.20 & 0.45 & 0.30 & 0.05  \\
DKTD & 50k & 6 & 0.15 & 0.35 & 0.35 & 0.15 \\
DKTD & 75k & $\infty$ & 0.15 & 0.30 & 0.30 & 0.25 \\
\midrule
Meta-World & 0 & 2 & 0.30 & 0.55 & 0.15 & 0.00 \\
Meta-World & 25k & 4 & 0.20 & 0.45 & 0.30 & 0.05  \\
Meta-World & 50k & 6 & 0.15 & 0.35 & 0.35 & 0.15 \\
Meta-World & 75k & $\infty$ & 0.15 & 0.30 & 0.30 & 0.25  \\
\bottomrule
\end{tabular}
\end{table}

\FloatBarrier
\section{Extended Experiments}
\label{appendix:ablation}

\begin{figure}[h]
    \centering
    \includegraphics[width=0.7\linewidth]{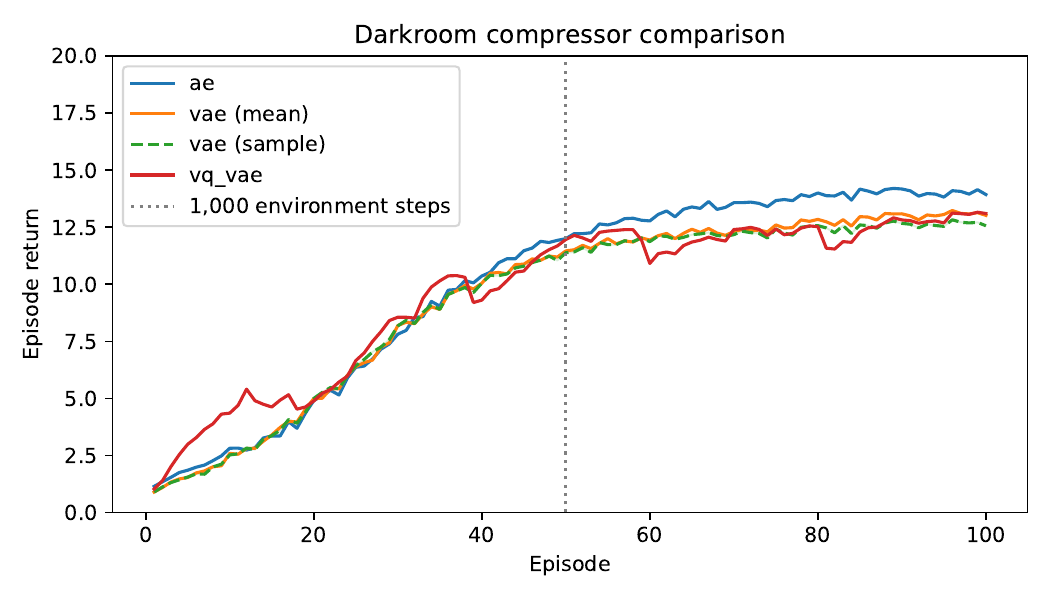}
    \caption{Compressor ablation in Darkroom. Episode returns for RAD with AE, VAE (posterior mean or sampling), and VQ-VAE compressors. AE achieves the highest late-stage returns. The vertical dotted line marks the 1,000-step training-history horizon (50 episodes).}
    \label{fig:compressor-ablation}
\end{figure}

\paragraph{Comparison of Different Compressors.}
We compare three Transformer-based compressors in Darkroom, using the same backbone, latent-token dimensions, and recurrent memory update:
\begin{itemize}
    \item \textbf{AE:} the compressor adopted in this work, featuring
    a deterministic autoencoder with continuous latent representations.
    \item \textbf{VAE:} a variational autoencoder with a Gaussian latent distribution and KL regularization, evaluated using either posterior means or sampled latents \citep{kingma2013auto}.
    \item \textbf{VQ-VAE:} a vector-quantized autoencoder that maps compressor outputs to a learned discrete codebook before the recurrent memory update \citep{van2017neural}.
\end{itemize}
Shown in Figure \ref{fig:compressor-ablation}, all variants improve with accumulated interaction. Its advantage becomes clearer beyond 50 episodes, corresponding to the 1,000-step training-history horizon. VAE performs similarly with posterior means and sampled latents, with a small late-stage advantage for mean-based evaluation. VQ-VAE initially improves faster but exhibits larger fluctuations and ultimately attains returns comparable to VAE. These results favor the deterministic continuous AE for this setting: neither variational regularization nor vector quantization improves downstream performance.

\begin{figure}
    \centering
    \includegraphics[width=0.6\linewidth]{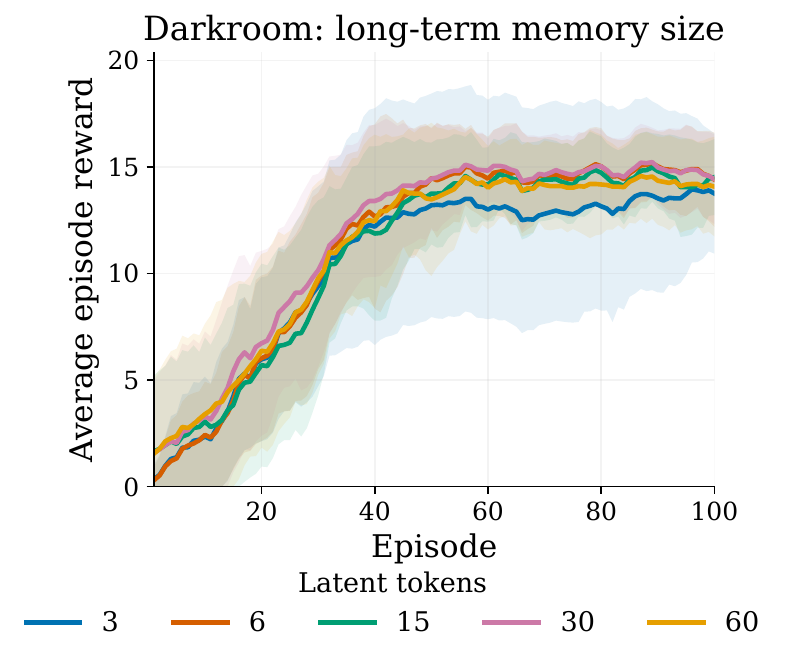}
    \caption{Average rewards of RAD with different long-term memory size on Darkroom}
    \label{fig:memory-size-ablation}
\end{figure}

\paragraph{Effect of the Long-term memory size.}

Figure \ref{fig:memory-size-ablation} shows the performance of RAD with the same working memory size and different long-term memory size. RAD successfully exhibits in-context adaptation even for a small long-term memory, possibly since preserving information in Darkroom is relatively simple. A long-term memory with 15 latent tokens exhibits the best performance while maintaining a relatively small long-term memory.

\begin{figure}
    \centering
    \includegraphics[width=0.6\linewidth]{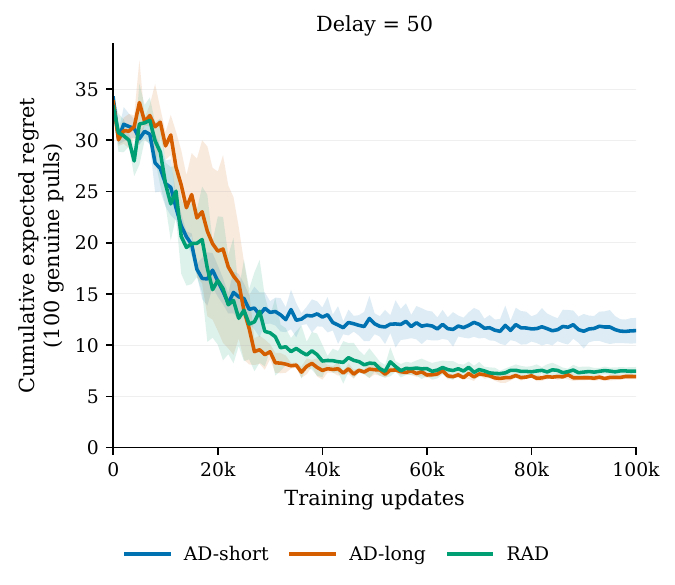}
    \caption{Training convergence of AD-short, AD-long, and RAD under delayed feedback (delay = 50): mean cumulative expected regret over 100 genuine pulls (50 before and 50 after the delay) on 100 fixed held-out tasks, evaluated periodically during distillation, with shading showing one standard deviation across 3 training seeds; RAD starts from compression-pretrained checkpoints (pretraining updates excluded from the horizontal axis), and all methods receive the same distillation update budget.}
    \label{fig:convergence}
\end{figure}

\paragraph{Convergence speed of AD/RAD.} Figure~\ref{fig:convergence} tracks online regret on held-out tasks as a function of the number of training updates; all three variants start from a regret of $\approx$33 at initialization and improve steadily thereafter. AD-short descends fastest early but saturates at a suboptimal plateau, since its short context cannot fully resolve feedback that is 50 steps old; AD-long, whose context spans the entire history, learns more slowly---it still trails AD-short at 20k updates---but does not saturate, crossing below AD-short around 25--30k updates and converging to $6.9$. RAD converges the slowest among the methods, achieving the optimal performance at around 60k updates, yet it is more efficient during inference, matching AD-long's performance with significantly smaller context window.

\section{Extended Related Work}

\paragraph{History and Trajectory Representations in RL.}
RAD is also related to methods that learn compact representations of
trajectories or interaction histories.
SeCTAR \citep{co2018self} learns latent trajectory representations for
hierarchical RL, while history-based representation methods study how past
observations can be summarized into compact states for control
\citep{patil2024learning,ni2024bridging}.
HELM \citep{paischer2022history} similarly constructs compact history representations
for partially observable RL.
Trajectory simplification has also been formulated as an RL problem
\citep{wang2021trajectory}, where redundant trajectory points are removed while
preserving geometric information.
These methods primarily seek representations sufficient for control,
prediction, or trajectory reconstruction.
RAD instead targets the learning history of an in-context RL agent:
its long-term memory recursively summarizes experience across episodes and
is optimized to retain the information needed to continue the adaptation
process of the distilled source algorithm.

\paragraph{Meta Reinforcement Learning.} 

Meta-reinforcement learning (Meta-RL) aims to learn the underlying structure of RL, enabling rapid adaptation to new tasks. Early deep meta-RL methods primarily focused on online adaptation, where recurrent architectures are trained to implicitly implement RL algorithms through their internal states, enabling fast within-episode learning \citep{duan2016rl2fastreinforcementlearning, wang2017learningreinforcementlearn}. On the other hand, gradient-based meta-learning approaches seek to learn parameter initializations that can be efficiently adapted to new tasks via a small number of gradient updates \citep{finn2017modelagnostic,nichol2018firstordermetalearningalgorithms}.

More recently, offline meta-RL has emerged as an important paradigm by leveraging pre-collected datasets of trajectories to meta-train without additional environment interaction. Existing offline meta-RL methods span multiple classes, including gradient-based adaptation \citep{mitchell2021offlinemeta}, Bayesian inference over latent task variables \citep{rakelly2019efficientoffpolicy, dorfman2021offlinemeta}, and representation-learning or contrastive approaches for task inference from offline data \citep{yuan2022robusttask}.

\end{document}

%% file: math_commands.tex
\usepackage{amsmath,amsfonts,bm}

\def\eqref#1{equation~\ref{#1}}

\def\1{\bm{1}}

\DeclareMathAlphabet{\mathsfit}{\encodingdefault}{\sfdefault}{m}{sl}
\SetMathAlphabet{\mathsfit}{bold}{\encodingdefault}{\sfdefault}{bx}{n}

